\documentclass{bmvc2k}

\newcommand{\bm}[1]{\mathbf{#1}} %Bold vectors and matrices

\newcommand\raiseT[2]{%
\setbox0\hbox{$#1{#2}$}\raise\dp0\box0}

\usepackage{amsmath,amssymb}
\usepackage{booktabs}

\title{Fill in Fabrics: Body-Aware Self-Supervised Inpainting for Image-Based Virtual Try-On}

\addauthor{Hasib Zunair}{hasibzunair@gmail.com}{1}
\addauthor{Yan Gobeil}{yan.gobeil@decathlon.com}{2}
\addauthor{Samuel Mercier}{samuel.mercier@decathlon.com}{2}
\addauthor{A. Ben Hamza}{hamza@ciise.concordia.ca}{1}

\addinstitution{
 Concordia University \\
 Montreal, QC, Canada
}

\addinstitution{
Decathlon Canada \\
Montreal, QC, Canada
}

\runninghead{Zunair, Hamza}{Self-Supervised Inpainting for Virtual Try-On}

\begin{document}

\maketitle

\begin{abstract}
Previous virtual try-on methods usually focus on aligning a clothing item with a person, limiting their ability to exploit the complex pose, shape and skin color of the person, as well as the overall structure of the clothing, which is vital to photo-realistic virtual try-on. To address this potential weakness, we propose a fill in fabrics (FIFA) model, a self-supervised conditional generative adversarial network based framework comprised of a Fabricator and a unified virtual try-on pipeline with a Segmenter, Warper and Fuser. The Fabricator aims to reconstruct the clothing image when provided with a masked clothing as input, and learns the overall structure of the clothing by filling in fabrics. A virtual try-on pipeline is then trained by transferring the learned representations from the Fabricator to Warper in an effort to warp and refine the target clothing. We also propose to use a multi-scale structural constraint to enforce global context at multiple scales while warping the target clothing to better fit the pose and shape of the person. Extensive experiments demonstrate that our FIFA model achieves state-of-the-art results on the standard VITON dataset for virtual try-on of clothing items, and is shown to be effective at handling complex poses and retaining the texture and embroidery of the clothing.
\end{abstract}

%-------------------------------------------------------------------------

\section{Introduction} \label{intro}
The core objective of image-based virtual try-on is to synthesize a person image with a new clothing, given the image of the person wearing a different clothing item and the new clothing item as inputs. Virtual try-on can be broken down into three main sub-tasks, namely image warping, image compositing, and synthesizing. The latter is very challenging as a synthetic image must preserve the person's identity, pose and shape. Also, the occluded body parts in a clothing item should be correctly synthesized. Moreover, the clothing image should accurately fit the pose and shape of a person, and the details of the clothing should also be preserved (i.e. logo, texture and embroidery). Prior work~\cite{jetchev2017conditional,han2018viton,wang2018toward,yu2019vtnfp,han2019clothflow,minar2020cp,jandial2020sievenet,yang2020towards,yang2020towards,ge2021disentangled,ren2021cloth} formulates virtual try-on as a supervised learning problem by following two major steps: warp the clothing image to fit the human body/shape and fuse the warped clothing with the person image (i.e. compositing and synthesis). While most of these methods are able to preserve the identity of a person, there exists a significant gap towards photo-realism as they tend to fail not only in cases of complex pose and shape of the person, but also in synthesizing initially occluded body parts (e.g., long sleeve clothing). These methods also fail to preserve the logo, texture and embroidery of the clothing, as well as the overall shape of the clothing item. This is largely attributed to the objective functions used in the existing virtual try-on methods. In fact, many approaches use per-pixel-based, perceptual-based losses~\cite{wang2018toward,minar2020cp,yang2020towards,choi2021viton,jandial2020sievenet,ren2021cloth} and adversarial losses~\cite{jetchev2017conditional,ge2021disentangled}, which do not enforce any global context and semantics necessary to accurately model the human and clothing interaction for compositing and synthesis. In addition, existing virtual try-on methods~\cite{jetchev2017conditional,han2018viton,yu2019vtnfp,wang2018toward,minar2020cp,yang2020towards,choi2021viton,jandial2020sievenet,ren2021cloth} do not provide robustness performance for in-the-wild images. Therefore, it remains an open question as to how these methods would generalize in-the-wild and it is of paramount importance to develop methods that can overcome these challenges for highly photo-realistic virtual try-on.

In order to address the aforementioned limitations, we introduce a self-supervised conditional generative adversarial network model, dubbed Fill In FAbrics (FIFA), which is a body-aware inpainting framework for image-based virtual try-on. The proposed FIFA framework can synthesize more realistic logo, texture and embroidery of the target clothing and also tackles well person images with complex poses (e.g., hands occluded). Our approach consists of a Fabricator and a unified virtual try-on pipeline with a Segmenter, Warper and Fuser. The Fabricator is used as a form of self-supervised pretraining for Warper. The goal of the Fabricator is to reconstruct full clothing details, given a partial input, enabling the model to learn the overall structure of the clothing (i.e. logo, texture, embroidery, full/short sleeves). To enforce global context at multiple scales for accurate modeling of the human and clothing interaction for compositing and synthesis, we also propose to use a multi-scale structural constraint to warp and refine the target clothing. The main contributions of this paper can be summarized as follows:
\begin{itemize}
\item We propose FIFA, a self-supervised conditional generative adversarial network model for virtual try-on, which can handle the complex pose of a reference person while preserving the target clothing details.

\item We design a masked cloth modeling (MCM) objective to learn the overall structure of the clothing by predicting the full clothing image, given a masked input, for the downstream task of better target cloth warping and refinement.

\item We show through experimental results and ablation studies that our model achieves competitive performance in comparison with strong baselines, yielding more realistic virtual try-on outputs.
\end{itemize}

\section{Related Work} \label{related}
\noindent\textbf{Image-Based Virtual Try-On.}\quad The basic objective of image-based virtual try-on is to synthesize a photo-realistic new image by overlaying a desired product image seamlessly onto the corresponding region of a clothed person. To achieve this goal, various image-based virtual try-on methods based on generative models have been proposed, Conditional Analogy Generative Adversarial Network (CA-GAN)~\cite{jetchev2017conditional}, Virtual Try-On Network (VITON)~\cite{han2018viton}, Characteristic-Preserving Virtual Try-On (CP-VTON) network~\cite{wang2018toward}, CP-VTON+~\cite{minar2020cp}, Disentangled Cycle-consistency Try-On Network (DCTON)~\cite{ge2021disentangled}, ClothFlow~\cite{han2019clothflow}, SieveNet~\cite{jandial2020sievenet}, Adaptive Content Generating and Preserving Network (ACGPN)~\cite{yang2020towards}, and Cloth Interactive Transformer (CIT)~\cite{ren2021cloth}. While these methods aim to handle complex textures on clothes and reduce artifacts in the final try-on results, they fail when the visual difference between the person image and target clothing is significant (e.g., changing long sleeve clothing items with short sleeve) and also tend to generate distorted arm regions. Furthermore, they fail to tackle person images with complex poses.

\medskip\noindent\textbf{Masked Data Modeling.}\quad Masked data modeling has proven effective in natural language processing and computer vision~\cite{mikolov2013efficient,devlin2018bert,liu2018image,suvorov2022resolution}. Existing masked data modeling approaches include Context Encoders~\cite{pathak2016context} and Masked Autoencoders (MAE)~\cite{he2021masked}. Our work differs from existing methods in two main aspects. First, we make predictions at a pixel level compared to predicting visual tokens~\cite{bao2021beit}. Second, our encoder network is purely convolutional by design, and is not based on vision transformers, which have been shown to perform well only when pre-trained on large-scale image datasets such as the JFT-300M dataset~\cite{dosovitskiy2020vit}.

\section{Proposed Method} \label{method}
\noindent\textbf{Problem Statement.}\quad Image-based virtual try-on aims to synthetically fit a target clothing onto a reference person while preserving photo-realistic details such as identity, pose and shape of the person, as well as texture and embroidery of the target clothing. More precisely, given a reference person image and a clothing image, the goal of our proposed FIFA model is to synthesize a new image of the same person wearing the target clothing such that the shape and pose of the person, as well as the details of the clothing are preserved.

\subsection{Fill in Fabrics for Virtual Try-On}
\begin{figure}[!htb]
\centering
\includegraphics[scale=.58]{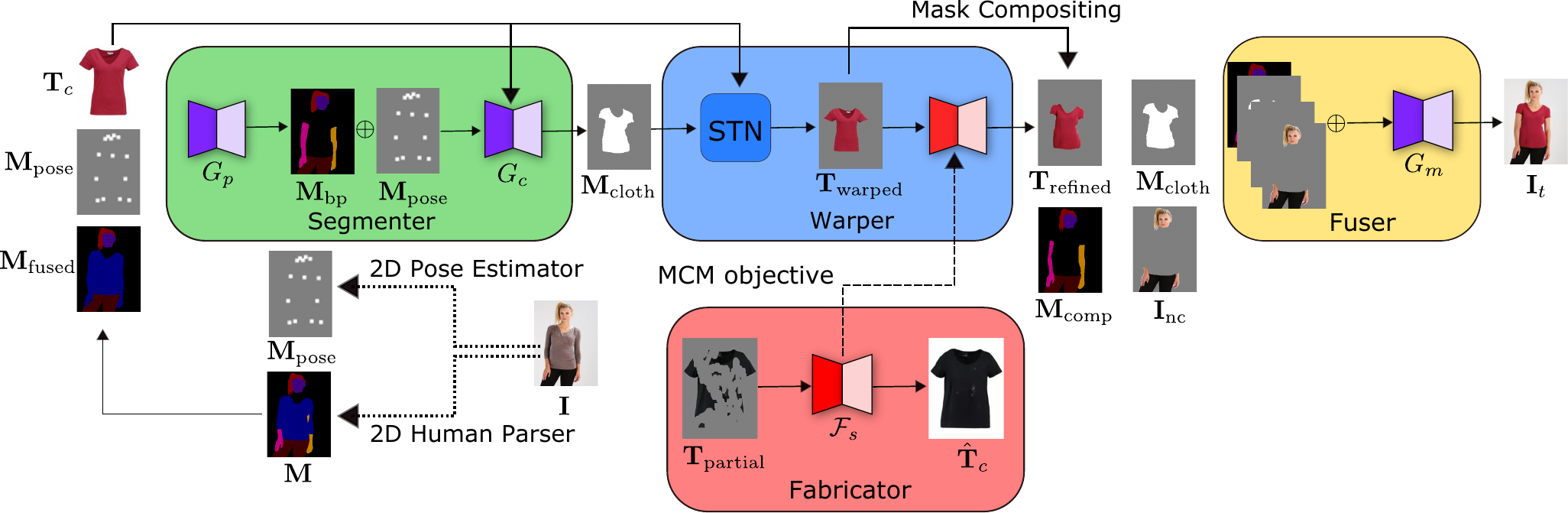}
\caption{Schematic layout of the proposed FIFA framework for virtual try-on. Given a person image $\bm{I}$ and a clothing image $\bm{T}_{c}$, FIFA synthesizes a try-on image $\bm{I}_{t}$, where the person in image $\bm{I}$ is wearing the target clothing $\bm{T}_{c}$. STN refers to the spatial transformer network, and $\oplus$ denotes concatenation.}
\label{fig:fifa}
\end{figure}

The proposed FIFA framework consists of a Fabricator and a unified pipeline consisting of a Segmenter, Warper and Fuser for virtual try-on, as shown in Figure~\ref{fig:fifa}. Given a partial input, we first use the Fabricator to reconstruct the full clothing details and learn the overall structure of the clothing (i.e. texture, full and half sleeve). This is used as a pretext task for the Warper. Second, we use Segmenter to predict the mask of the body parts of the reference person, as well as the masked target clothing regions. Third, we employ Warper to warp the target clothing image such that it fits the masked clothing region with the aim to capture the pose and shape of the reference person. Finally, Fuser integrates the outputs from Segmenter and Warper in order to synthesize the final try-on image.

\medskip\noindent\textbf{Fabricator.}\quad The Fabricator aims to reconstruct (i.e. fill in fabrics) the full target clothing image $\hat{\bm{T}}_{c}$, given the partial target clothing $\bm{T}_\text{partial}$. To this end, the Fabricator learns to represent the overall structure of the clothing while reconstructing the missing regions (i.e. fill in correct pixels that make sense in the context). Inspired by the concept of image inpainting (i.e. the task of filling in holes in an image) using partial convolutions, where the convolution is masked and re-normalized to be conditioned on only valid pixels~\cite{liu2018image}, we construct $\bm{T}_\text{partial}$ from $\bm{T}_{c}$ using masks of random streaks and holes of arbitrary shapes. In contrast to image inpainting, we formulate our objective as a masked cloth modeling problem, which can be regarded as a form of self-supervised pre-training for the downstream task of virtual try-on. More specifically, we train an encoder-decoder network $\mathcal{F}_s$ to reconstruct the reconstructed target clothing $\hat{\bm{T}}_{c}$, for a given $\bm{T}_\text{partial}$, with the goal to be close to the original target clothing image $\bm{T}_{c}$ (i.e. non-masked clothing) by minimizing the $L_1$ error $\mathcal{E}=\Vert\hat{\bm{T}}_{c}-\bm{T}_{c}\Vert_{1}$.

\medskip\noindent\textbf{Segmenter.}\quad The goal of the Segmenter is to preserve the body parts of the person during the synthesis process and also to accurately predict the semantic layout of the target clothing regions that are necessary for the Warper. Given a reference person image $\bm{I}$ and its associated mask $\bm{M}$ obtained via a publicly available human parser~\cite{li2020self}, the arms and torso regions are merged to form a fused map $\bm{M}_\text{fused}$. A conditional generative adversarial network (CGAN) $G_{p}$ is then trained to generate a different person body part mask $\bm{M}_\text{bp}$, which is conditioned on $\bm{M}_\text{fused}$, the 18-keypoint pose heatmap $\bm{M}_\text{pose}$ using out-of-the-box 2D pose estimator~\cite{8765346,cao2017realtime}, and the target clothing image $\bm{T}_{c}$. To generate the target clothing region $\bm{M}_\text{cloth}$, another CGAN $G_{c}$ is trained by combining $\bm{M}_\text{bp}$, $\bm{M}_\text{pose}$ and $\bm{T}_{c}$. Hence, in the Segmenter there are two CGANs in which the discriminator is similar to pix2pixHD~\cite{wang2018high} and the generator is a Residual U-Net architecture~\cite{zhang2018road} built on top of the U-Net model~\cite{ronneberger2015u} with residual connections~\cite{he2016deep}. This not only helps retain fine-grained features and predict accurate body part masks, but also helps generate better try-on results. For a given CGAN (i.e. $G_{p}$ or $G_{c}$), the adversarial loss is given by
\begin{equation}
\begin{split}
\mathcal{L}_\text{CGAN} = \mathbb{E}_{\bm{x}\sim p_{\text{data}}(\bm{x})}[\log D(\bm{x}\vert\bm{y})] +
\mathbb{E}_{\bm{z}\sim p_{z}(\bm{z})}[\log( 1 - D(G(\bm{z}\vert\bm{y})))],
\end{split}
\label{eq:advloss}
\end{equation}
where $G$ and $D$ are the generator and discriminator, $\bm{x}$ and $\bm{y}$ are the input and ground-truth mask, and $\bm{z}$ is a noise prior drawn from a standard normal distribution. A CGAN is a type of GAN that takes advantage of auxiliary information during the training process. To train a CGAN, we train the generator and discriminator simultaneously to maximize the performance of both. In simple terms, the goal of the generator is to generate data that the discriminator classifies as ``real'', whereas the objective of the discriminator is to not be ``fooled'' by the generator. In other words, the generator and discriminator follow the two-player min-max game with $\mathcal{L}_\text{CGAN}$ as a function of $G$ and $D$.

In order to enforce consistency at the pixel-level, we also use the pixel-wise cross-entropy loss $\mathcal{L}_\text{CE}$ for better semantic segmentation results from the generator. Therefore, the overall objective is defined as
\begin{equation}
\mathcal{L}_\text{mask} = \alpha_{1} \mathcal{L}_\text{CGAN}  + \alpha_{2} \mathcal{L}_\text{CE},
\label{eq:seg_loss}
\end{equation}
where $\alpha_{1}$ and $\alpha_{2}$ are nonnegative regularization parameters, which control the contribution of each loss term. Following previous work~\cite{yang2020towards}, we set $\alpha_{1}$ and $\alpha_{2}$ to 1 and 10, respectively, in our experiments.

\medskip\noindent\textbf{Warper.}\quad We employ Warper to naturally deform the target clothing to fit the mask of the clothing region with respect to the pose of the person, as well as to preserve the texture and embroidery of the target clothing. While the Adaptive Content Generating and Preserving Network (ACGPN)~\cite{yang2020towards} for virtual try-on has been shown effective at predicting the semantic layout of the reference image, it fails, however, to preserve complex poses, logo, texture and embroidery of the target clothing. This is largely due to the fact that ACGPN employs the Spatial Transformer Network (STN)~\cite{jaderberg2015spatial} with Thin Plate Splines (TPS)~\cite{duchon1977splines} and an additional refinement network U-Net~\cite{ronneberger2015u}. To address these limitations, we design a masked cloth modeling objective (MCM) when training the Warper to better preserve logo, texture and embroidery of the target clothing. More specifically, we transfer the learned representations in $\mathcal{F}_{s}$ from Fabricator to the refinement network. We also incorporate a multi-scale structural constraint (MSC) to enforce global context at multiple scales for better warping of the target clothing according to the pose and shape of the person. Our strategy of training Warper yields better warped target clothes, which have fine details (i.e. logo, texture and embroidery), and is especially effective at handling complex poses.

Given the target clothing region $\bm{M}_\text{cloth}$ and target clothing image $\bm{T}_{c}$, the goal of Warper is to deform $\bm{T}_{c}$ such that it fits $\bm{M}_\text{cloth}$. STN first warps the clothing to $\bm{T}_\text{warped}$. This is further refined using $\bm{T}_\text{warped}$ as input to the refinement network with the goal to generate more details (i.e. logo, texture, embroidery). In a similar vein to~\cite{wang2018toward,yang2020towards}, composition is then performed on the output of the refinement network with $\bm{M}_\text{cloth}$ to output the final refined clothing $\bm{T}_\text{refined}$. The overall loss for the STN in Warper is an unweighted combination of the $\mathcal{L}_\text{CGAN}$ loss and a second-order difference constraint~\cite{yang2020towards}. The losses for the refinement network (i.e. pre-trained from the encoder-decoder network $\mathcal{F}_{s}$) are $\mathcal{L}_\text{CGAN}$ and the perceptual $\mathcal{L}_\text{VGG}$ loss~\cite{johnson2016perceptual}. This VGG perceptual loss helps ensure the target clothing and its warped version contain the same semantic content. In addition, we introduce a multi-scale structural constraint to enforce global context at multiple scales during training. Therefore, the overall loss function is defined as
\begin{equation}
\mathcal{L}_\text{refined} = \beta_{1} \mathcal{L}_\text{CGAN}  + \beta_{2} \mathcal{L}_\text{VGG} + \beta_{3} \mathcal{L}_\text{MS-SSIM}
\label{eq:warp_loss}
\end{equation}
where $\beta_{1}$, $\beta_{2}$ and $\beta_{3}$ are regularization parameters, which are set to 0.2, 20 and 15, respectively, in our experiments. $\mathcal{L}_\text{MS-SSIM}$ is the multi-scale structural similarity constraint~\cite{zhao2016loss}. The Warper benefits from the MCM objective and is able to better preserve the logo, texture and embroidery of the target clothing. It also benefits from MSC to enforce global context in order to ensure better warping of the target clothing according to the pose and shape of the person. This in turn helps produce improved try-on results in Fuser.

\medskip\noindent\textbf{Fuser.}\quad The Fuser merges the target clothing region, refined clothing image, a composited body part mask and a body part image with original clothing region masked out in order to produce the final try-on image. First, the Fuser generates a composited body part mask to remove or preserve the non-target body parts, which correspond, in most cases, to the arms of the person. This is then used in the second stage to determine which parts to preserve or generate when synthesizing the final try-on results. Given the original body part mask $\bm{M}_\text{obp}$, the clothing mask $\bm{M}_\text{oc}$ from $\bm{M}$ (i.e. head, arms, torso removed), $\bm{M}_\text{bp}$ and $\bm{M}_\text{cloth}$ from Segmenter, the composited body part mask $\bm{M}_\text{comp}$ is given by
\begin{equation}
\bm{M}_\text{comp} = ((\bm{M}_\text{bp} \odot \bm{M}_\text{oc}) + \bm{M}_\text{obp}) \odot (\bm{J} - \bm{M}_\text{cloth}),
\label{eq:composition1}
\end{equation}
where $\odot$ denotes element-wise multiplication and $\bm{J}$ is an all-ones matrix. As this step takes an input from Segmenter, it is crucial to produce accurate segmentation maps of $\bm{M}_\text{bp}$ and $\bm{M}_\text{cloth}$ for better compositing. We also perform compositing on $\bm{I}$ to get the body part image with $\bm{I}_\text{nc}$ being the original clothing region masked out as follows:
\begin{equation}
\begin{split}
\bm{I}_\text{nc} &= (\bm{I} - \bm{M}_\text{oc}) \odot (\bm{J} - \bm{M}_\text{cloth}).
\end{split}
\label{eq:composition2}
\end{equation}
Hence, given $\bm{T}_\text{refined}$ from Warper, $\bm{M}_\text{cloth}$ from Segmenter, $\bm{M}_\text{comp}$ and $\bm{I}_\text{nc}$, we train a CGAN $G_{m}$ to predict the final try-on image $\bm{I}_{t}$ by minimizing the following loss function
\begin{equation}
\mathcal{L}_\text{fuser} = \gamma_{1} \mathcal{L}_\text{CGAN}  + \gamma_{2} \mathcal{L}_\text{VGG},
\label{eq:fuser_loss}
\end{equation}
where the hyper-parameters $\gamma_{1}$ and $\gamma_{2}$ are set to 1 and 10, respectively, in our experiments.

\section{Experiments} \label{experiments}
We conduct extensive experiments to assess the performance of the proposed FIFA framework in comparison with competing baseline models for virtual try-on. Experimental details and additional results and ablation studies are provided in the supplementary material. Code is available at: \textcolor{blue}{https://github.com/hasibzunair/fifa-tryon}

\subsection{Experimental Setup}
\noindent\textbf{Datasets.}\quad We demonstrate and analyze the performance of our model on two virtual try-on datasets: VITON and DecaWVTON.
\begin{itemize}
\item \textbf{VITON:} This dataset consists of 16,253 pairs of front-view women images and front-view top clothing images split into a training set of 14,221 pairs and a test set of 2,032 pairs. To evaluate the capability of virtual try-on methods in handling different poses of a person, we divide the VITON test set into three subsets of easy, medium and hard cases according to the human pose in the reference images. These test subsets are denoted as VITON-E, VITON-M and VITON-H for easy, medium and hard, respectively~\cite{yang2020towards}.

\item \textbf{DecaWVTON:} To demonstrate the generalizability of FIFA to in-the-wild images, we use DecaWVTON, a proprietary dataset comprised of images with complex poses and clothing not present in the VITON dataset (e.g., turtle neck). Also, the clothing images are rotated, whereas VITON consists of only front-view clothing images. In many cases, the head portion is cut out (i.e. either fully or partially), whereas in VITON the person images consist of full faces.
\end{itemize}

\medskip\noindent\textbf{Baselines.}\quad We evaluate the performance of our proposed virtual try-on model against recent state-of-the-art techniques, including CA-GAN~\cite{jetchev2017conditional}, VITON~\cite{han2018viton}, CP-VTON~\cite{wang2018toward}, CP-VTON+~\cite{minar2020cp}, SieveNet~\cite{jandial2020sievenet}, and segmentation based methods such as VTNFP~\cite{yu2019vtnfp} and ACGPN~\cite{yang2020towards}, as well as flow based methods such as ClothFlow~\cite{han2019clothflow}. We also compare our model against a cycle-consistency based approach DCTON~\cite{ge2021disentangled} and a transformer based method CIT~\cite{ren2021cloth}.

\medskip\noindent\textbf{Evaluation Metrics.}\quad Following previous work~\cite{minar2020cp,yang2020towards}, we use the Structural SIMilarity (SSIM) that captures image level similarity and the Frechet Inception Distance (FID) that captures the distributional similarity. Both metrics are commonly used for benchmarking virtual try-on methods to quantify the visual difference between the generated and real reference images. Higher scores of SSIM and lower scores of FID indicate higher quality of the synthesized results. It is important to mention that while computing the SSIM and FID metrics, the target clothing items are the same as in the reference person as it is not possible to acquire ground truth images for try-on results.

\medskip\noindent\textbf{Implementation Details.}\quad All experiments are performed on a Linux workstation running 4.8Hz, 64GB RAM and a single NVIDIA RTX 3080 GPU. Experiments are conducted using Python programming language and PyTorch deep learning framework. A full training of FIFA, along with the Fabricator on the VITON dataset, takes roughly seven days. During training, the target clothing item is the same as the one in the reference person image, as it is not possible to acquire triplets to compute the loss with respect to the ground truth.

\subsection{Qualitative Results}
\begin{figure}[!htb]
\centering
\includegraphics[height=4.8cm]{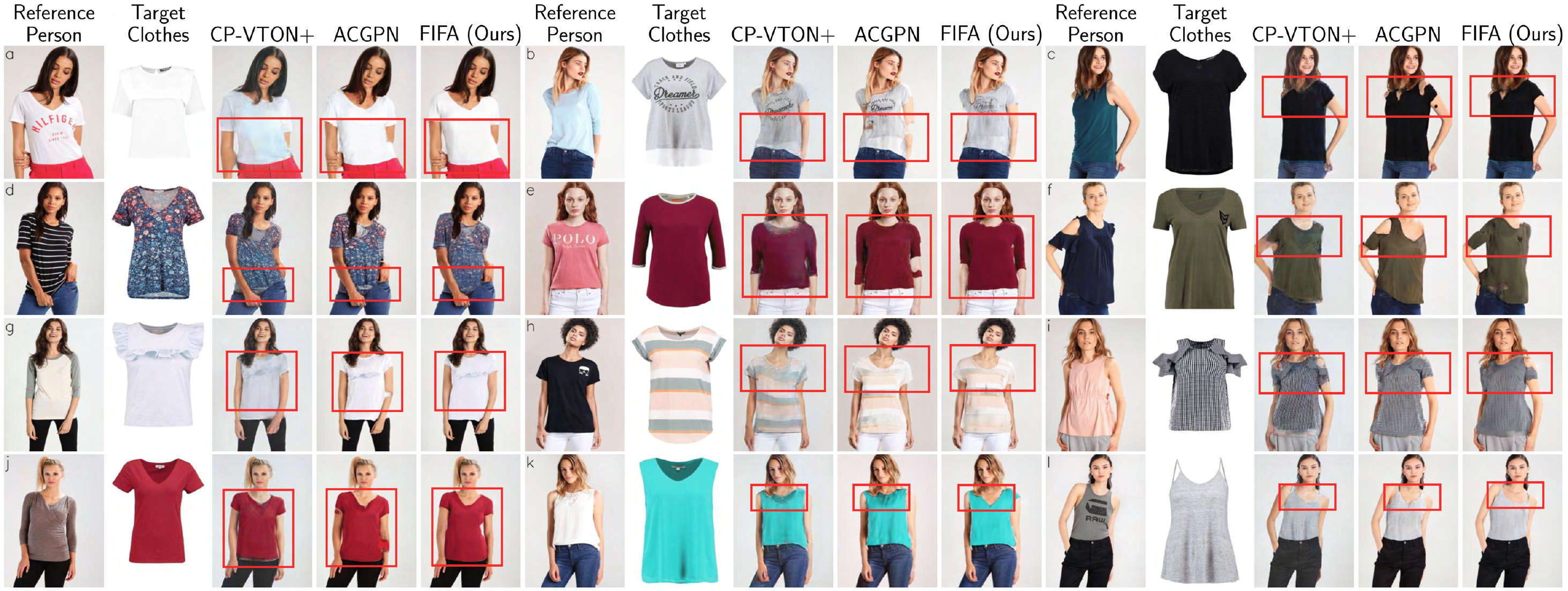}
\caption{Given a pair of a reference person image and a target clothing image, our FIFA model successfully synthesizes virtual try-on images. Compared to the baselines, FIFA is able to better handle complex poses and also retains photo-realistic details such as logo, texture, embroidery and structure (e.g., collar shape) of the target clothing.}
\label{fig:sota_vis}
\end{figure}

In Figure~\ref{fig:sota_vis}, we visually compare the performance of our proposed model with CP-VTON+ and ACGPN, which are state-of-the-art virtual try-on baselines. Each row shows a person virtually trying on different clothing items. As can be seen in the first row of Figure~\ref{fig:sota_vis}, when the pose of the reference person is complex (i.e. standing with arms behind the body), the baseline models either remove body regions, fail to warp short sleeve shirt, or add unrealistic body parts. These baselines are also unable to capture the global structure and semantics, which are needed for warping short sleeve shirts when the reference person is wearing a long sleeve shirt. This is due, in large part, to the limited capability of the warping strategies used in these baselines. The second and third rows of Figure~\ref{fig:sota_vis} show cases where the target clothing items are of complex texture (i.e. printed patterns, long sleeve, and shirt with logo) and embroidery (i.e. stripes). In these cases, CP-VTON+ fails to distinguish between the front and back part of the clothing regions, does not preserve the logo of the target clothes, and yields blurry results at the clothing and person body boundaries. While ACGPN produces non-blurry results, it fails to preserve the complex embroidery of the target clothing, and does not accurately warp long sleeve target clothing items, resulting in incomplete sleeves. In the last row of Figure~\ref{fig:sota_vis}, we can observe artifacts and mix-up of front and back part of the clothing in the images generated by the baseline methods. Also, both CP-VTON+ and ACGPN fail to capture the v-shaped structure of the target clothing, and do not accurately warp tank tops with very thin straps, resulting in either blurry or distorted clothing structure. Overall, these baselines fail to preserve the complex pose of the reference person, the complex texture and embroidery of the target clothing, and also the complex clothing types.

By comparison, our FIFA method is able to warp the target clothing in the case of complex poses, and preserves well the body parts. It benefits from the synergy between the MCM objective and the MSC constraint, which help preserve the pose of a person, capture the fine details of the target clothing (i.e. logo and embroideries), as well as the global structure of the clothing (i.e. front and back part of clothing, v-shaped collar). Moreover, FIFA benefits from residual blocks (RBs) to better predict the semantic layout of the body parts, resulting in realistic try-on results. In summary, this helps not only preserve the logo, texture, embroidery and the type of target clothing, but also yields an output having less artifacts and retains clear body parts, achieving more realistic try-on results. We also find that FIFA is able to better preserves the skin color of the person and accurately synthesizes the person's body parts, which were initially occluded. In addition, it can distinguish between the front and back part of clothing items.

It is worth pointing out that some examples in Figure~\ref{fig:sota_vis} seem to have color mismatch between synthetic clothes and target. We hypothesize that this might be attributed to the Warper, which in some cases produces blurry target clothing outputs. While using the multi-scale structural constraint (MSC) in Warper can output fine details of clothing, we argue that designing perceptually motivated loss functions may further improve the results.

\subsection{Quantitative Results}
Table~\ref{table:sota} shows that FIFA consistently outperforms all baselines, achieving relative improvements of 4.85\%, 4.22\%, 4.64\% and 4.47\% over the strongest ACGPN baseline on all (VITON), easy, medium and hard cases in terms of the SSIM metric. FIFA also outperforms ACGPN with a substantial relative improvement of 19.11\% in terms of FID.

\begin{table}[!htb]
\setlength{\tabcolsep}{4pt}
\begin{center}
\caption{Performance comparison of FIFA and state-of-the-art methods on the VITON, VITON-E, VITON-M and VITON-H test sets using SSIM and FID scores. FIFA consistently outperforms the baselines across easy, medium and hard cases. Boldface numbers indicate the best performance, whereas the best baselines are underlined.}
\medskip
\label{table:sota}
\begin{tabular}{llllll}
\hline\noalign{\smallskip}
\multicolumn{3}{r}{\bf SSIM ($\uparrow$)}\\
\cline{2-5}
\noalign{\smallskip}
\bf Method & VITON & VITON-E & VITON-M & VITON-H & \bf FID ($\downarrow$)\\
\noalign{\smallskip}
\hline
\noalign{\smallskip}
CA-GAN~\cite{jetchev2017conditional}  & 0.740 & - & - & - & 47.34 \\
VITON~\cite{han2018viton} & 0.783 & 0.787 & 0.779 & 0.779 & 55.71 \\
CP-VTON~\cite{wang2018toward}  & 0.745 & 0.753 & 0.742 & 0.729 & 24.43 \\
VTNFP~\cite{yu2019vtnfp} & 0.803 & 0.810 & 0.801 & 0.788 & - \\
ClothFlow~\cite{han2019clothflow}  & 0.843 & - & - & - & 23.68 \\
CP-VTON+~\cite{minar2020cp} & 0.750 & - & - & - & 21.08 \\
SieveNet~\cite{jandial2020sievenet}  & 0.837 & - & - & - & 26.67 \\
ACGPN~\cite{yang2020towards} & \underline{0.845} & \underline{0.854} & \underline{0.841} & \underline{0.828} & 16.64 \\
DCTON~\cite{ge2021disentangled}  & 0.830 & - & - & - & \underline{14.82} \\
CIT~\cite{ren2021cloth} & 0.827 & - & - & - & - \\ %\midrule
\bf FIFA (Ours)  & \textbf{0.886} & \textbf{0.890} & \textbf{0.880} & \textbf{0.865} & \textbf{13.46} \\
\hline
\end{tabular}
\end{center}
\end{table}

Interestingly, our FIFA model yields significant relative improvements of 5.10\% and 43.16\% over ClothFlow in terms of SSIM and FID, respectively. It is worth pointing out that ClothFlow operates on streams (i.e. optimal flow maps) to predict the movement of clothes and is computationally expensive, while our model is a purely image-based virtual try-on approach operating on image pixels. Our method also outperforms the transformer based CIT baseline with a relative improvement of 7.13\% in terms of SSIM. This better performance of our approach is significant because transformers are built on self-attention operations and are quite strong in modeling the global context between the person and target clothing. In addition, transformers in computer vision tasks perform well only when pre-trained on a large cohort of images such as the JFT-300M dataset, which is comprised of 18K classes and 303M high-resolution images~\cite{dosovitskiy2020vit}.

\subsection{Ablation Study}
\begin{figure}[!htb]
\centering
\includegraphics[height=4.65cm]{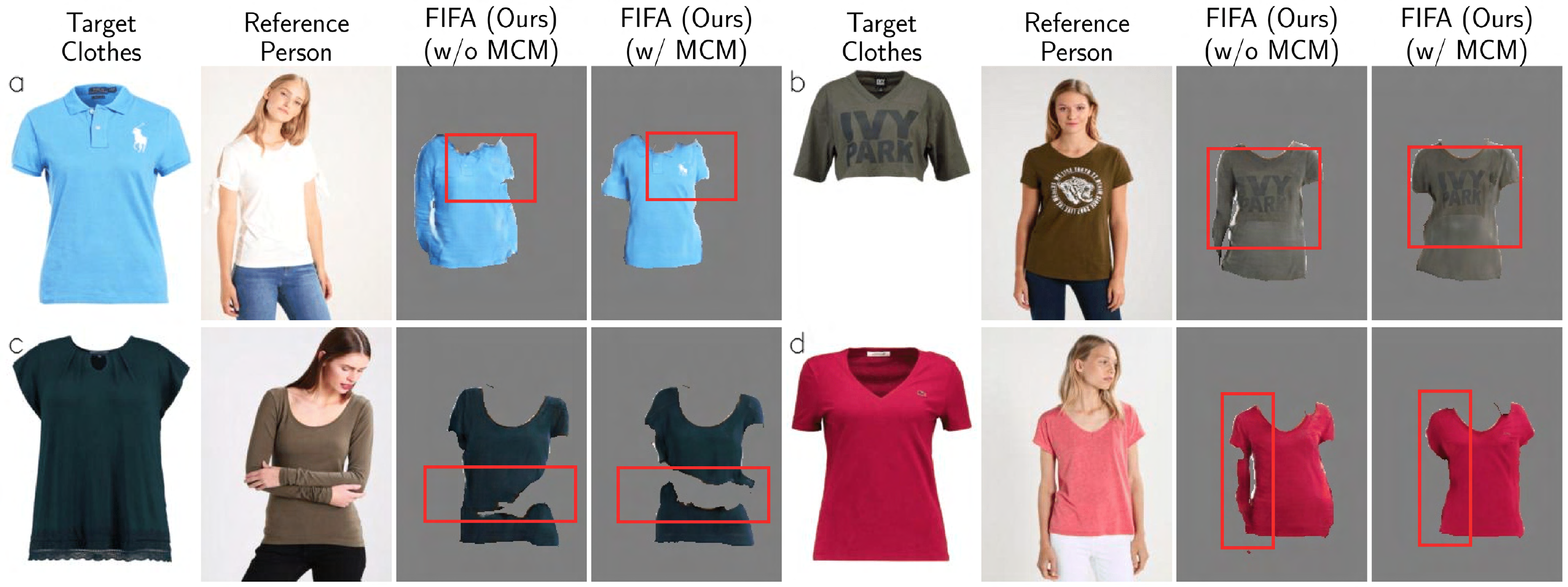}
\caption{Warped target clothing results, demonstrating the effectiveness of the MCM objective in Warper. Warper with MCM is capable of handling complex poses (i.e. body-aware) and preserving the logo and embroidery of the clothing.}
\label{fig:mcm_abl}
\end{figure}

\noindent\textbf{Effectiveness of Masked Cloth Modeling (MCM).}\quad Figure~\ref{fig:mcm_abl} illustrates the benefit of using the MCM objective in preserving the pose and logo, as well as in accurately warping the target clothing. As can be seen, MCM is able to preserve the logo of the target clothing, whereas without MCM the logo is completely lost. MCM also helps in accurately preserving or synthesizing body parts in complex poses (i.e. body aware), as well as in accurately warping the target clothing. Notice that without MCM, there is a problem of unnecessarily editing regions of the target clothing (i.e. making half sleeve shirt a full sleeve). This is largely attributed to the richer learning signal provided by MCM rather than just using the supervised objective of predicting the warped clothing, which fits the reference person, enabling our approach to accurately model the interactions between the target clothing and reference person clothing.

\begin{figure}[!htb]
\centering
\includegraphics[height=4.65cm]{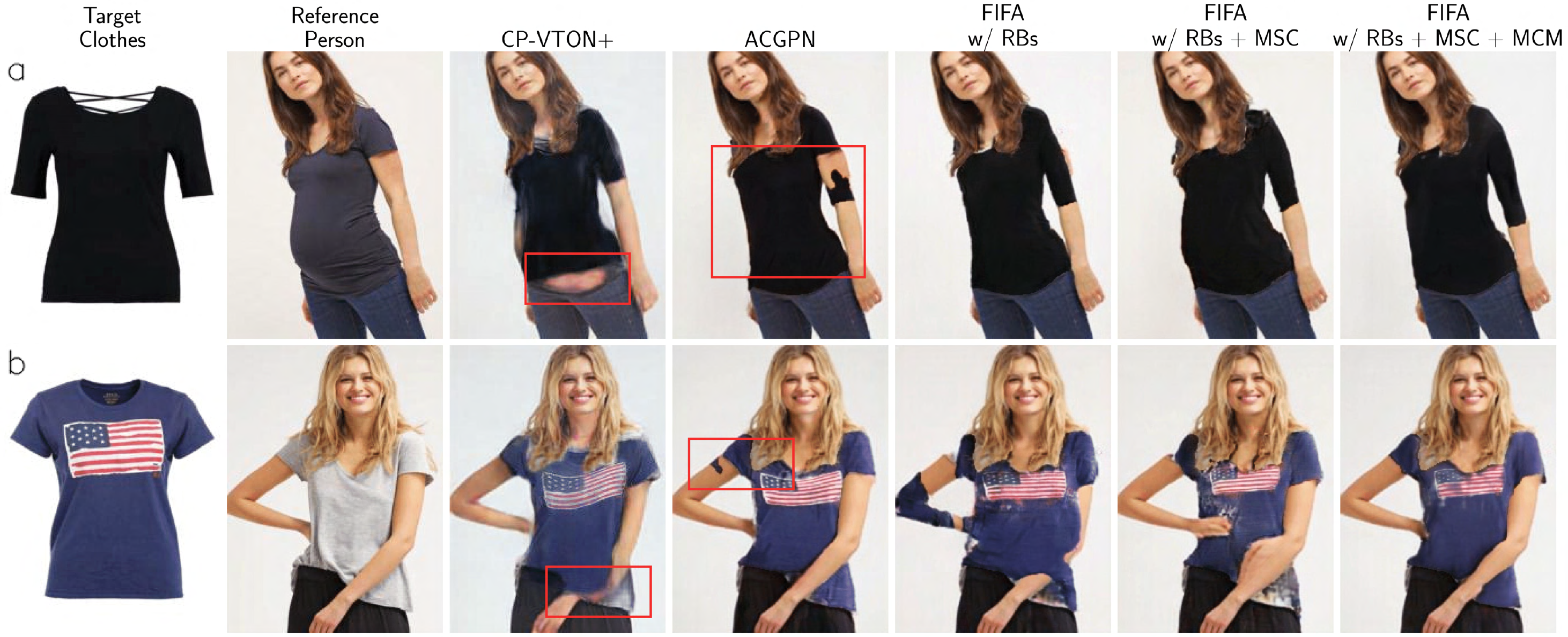}
\caption{Try-on results, demonstrating the effectiveness of MSC in Warper. Warper with MSC helps capture global context of the target clothing and preserves the shape of the person.}
\label{fig:msc_abl}
\end{figure}

\medskip\noindent\textbf{Effectiveness of Multi-Scale Structural Constraint (MSC).}\quad Figure~\ref{fig:msc_abl} shows that the use of per-pixel-based and perceptual-based loss functions~\cite{wang2018toward,minar2020cp,yang2020towards,choi2021viton,jandial2020sievenet,ren2021cloth} is not enough to capture the global context and semantics, which are needed for preserving the shape of a person and also for realistically synthesizing body parts. The per-pixel-based loss function $\mathcal{L}_1$ measures the distance between pixels and does not enforce any global constraint. On the other hand, the perceptual loss $\mathcal{L}_\text{VGG}$ quantifies the similarity between the reconstructed and ground-truth images, but only at a latent representation level (i.e. computes the distance of the features extracted by VGG-19~\cite{simonyan2014very}). Also, it tends to generate artifacts~\cite{johnson2016perceptual}, which is in line with our findings. We show that by adding MSC, our model is able to better tackle these issues and learns to exploit context at different scales, while the CP-VTON+ and ACGPN baselines introduce artifacts and do not preserve well the shape of the person.

\subsection{Generalization to In-The-Wild Virtual Try-On}
To test the generalizability of virtual try-on models to in-the-wild images, we set up a challenging task where the results would better reflect the robustness on unseen data. We compare FIFA against the state-of-the-art ACGPN model~\cite{yang2020towards} by training both methods on VITON and testing them on DecaWVTON. Results presented in the supplementary material demonstrate that FIFA yields substantial improvements over ACGPN in terms of SSIM and FID, indicating that FIFA is more robust to in-the-wild images for virtual try-on.

\section{Conclusion} \label{conclusion}
We introduced a body-aware self-supervised inpainting framework for image-based virtual try-on with a focus on tackling complex poses, learning the overall structure of clothing and incorporating global context. Our proposed FIFA model achieves significant improvements in the synthesized try-on image by not only retaining the logo, texture and embroidery of the clothing, but also able to better handle the complex poses, indicating that it is body aware, a crucial feature for photo-realistic virtual try-on. By combining the strengths of mask cloth modeling, multi-scale structural constraint and residual blocks, FIFA outperforms strong baselines on the VITON dataset across all, easy, medium and hard cases. In addition, we set up an evaluation framework for testing robustness of virtual try-on models to in-the-wild images and found that FIFA outperforms previous state-of-the-art methods by a significant margin.

\bibliographystyle{bmvc2k_natbib}
\bibliography{references}
\end{document}